\documentclass[runningheads]{llncs}
\usepackage[T1]{fontenc}
\usepackage{amssymb}
\usepackage{graphicx}
\usepackage{url}
\begin{document}
\title{
Model-Based Systems Engineering Framework for SysML-Driven Design of Autonomous UAVs
}
%
\titlerunning{MBSE-SysML-ROS 2 Framework}
%

\author{
Deekshitha Angadi\inst{1,2} \and
Naveena Budda\inst{3} \and
Vikas Agarwal\inst{4} \and
Mohamed Samshad\inst{5} \and
Bharath Kumar Suryadevara\inst{6} \and
Narsimlu Kemsaram\inst{7}
}
\authorrunning{D. Angadi et al.}
%
\institute{
AIR Lab, UAVs Group, Autonomous Robotics Systems Limited, Hyderabad, India \and
Department of Microelectronics and VLSI Design, University of Hyderabad, Hyderabad, India \and
Department of IoT, Ideabytes Software India Private Limited, Hyderabad, India \and
School of Computer Science, Georgia Institute of Technology, Atlanta, USA \and
Department of Electrical Engineering, Indian Institute of Technology, Kanpur, India \and
Department of System Engineering, Akkodis AS\&D GmbH, Bremen, Germany \and
Department of Artificial Intelligence, Universiti Malaya, Kuala Lumpur, Malaysia
}

\maketitle              
\begin{abstract}

Autonomous Unmanned Aerial Vehicles (UAVs) are complex cyber-physical systems that require the coordinated integration of flight control, navigation, perception, communication, power management, and mission-level decision-making under safety, timing, and reliability constraints. However, many autonomous UAV development workflows still rely on document-centric requirements, separated architectural descriptions, and software implementation artifacts, which can lead to ambiguity, interface inconsistencies, and weak traceability during early design. This paper presents a Model-Based Systems Engineering (MBSE) design framework for the SysML-driven development of autonomous UAVs. The proposed framework uses the Systems Modeling Language (SysML) as a formal design backbone to structure UAV development across four connected layers: stakeholder requirements, functional decomposition, logical architecture, and physical/software allocation. SysML requirement diagrams, activity diagrams, block definition diagrams, internal block diagrams, state machine diagrams, and parametric diagrams are used to capture the functional, structural, behavioral, interface, and performance aspects of the UAV system. The logical architecture is then systematically mapped to a Robot Operating System 2 (ROS 2) software architecture by relating SysML blocks to ROS 2 nodes, flow ports and connectors to topics, request-response interactions to services, and goal-oriented behaviors to actions. The framework is illustrated at the design level using representative autonomous UAV mission scenarios, including autonomous take-off, waypoint navigation, hover stabilization, obstacle avoidance, return-to-home, and emergency handling. The resulting model supports requirement allocation, interface definition, subsystem responsibility assignment, and verification planning before simulation or physical deployment. This design-phase study provides a traceable and reproducible methodology for reducing early-stage design ambiguity and preparing autonomous UAV systems for future Gazebo simulation, hardware-in-the-loop validation, and real-world flight testing.


\end{abstract}

\section{Introduction}



Autonomous UAVs are increasingly used in infrastructure inspection, surveillance, precision agriculture, environmental monitoring, emergency response, and other mission-critical aerial applications. Unlike remotely piloted aircraft, autonomous UAVs are expected to perceive the environment, estimate their own state, plan mission actions, execute flight behaviors, monitor system health, and respond to abnormal conditions with limited or no continuous human intervention. These capabilities require the coordinated integration of multiple cyber-physical subsystems, including sensing, perception, localization, mapping, flight control, navigation, communication, power management, and mission-level decision-making. As the number of subsystems, interfaces, and operating modes increases, the early design of autonomous UAVs becomes a major systems engineering challenge.
A common limitation in autonomous UAV development is the continued reliance on document-centric engineering practices. In such workflows, stakeholder requirements, subsystem descriptions, interface control documents, software architecture diagrams, and verification plans are often created and maintained separately. This separation can lead to ambiguous requirements, incomplete interface definitions, inconsistent behavioral logic, and weak traceability between stakeholder needs and implementation artifacts. These issues are particularly critical for autonomous UAVs because safety-related behaviors such as return-to-home, emergency handling, low-battery response, communication-link-loss handling, and obstacle avoidance must be clearly specified before simulation or physical flight testing. If these design dependencies are not captured during the early design phase, defects may only become visible during software integration, simulation, or real-world testing, where correction is more difficult and costly.

MBSE provides a structured approach for managing this complexity by using formal system models as the central representation of requirements, architecture, behavior, interfaces, constraints, and verification information \cite{incose2023,iso15288,madni2018}. Instead of treating requirements and architecture as disconnected documents, MBSE supports the creation of a connected model in which stakeholder needs can be traced to functional behaviors, subsystem responsibilities, interface flows, implementation elements, and verification activities. The SysML is widely used for MBSE because it provides standard diagram types for modeling requirements, structure, behavior, parametric constraints, and verification relationships \cite{friedenthal2014,delligatti2013,omg2019}. These capabilities make SysML well-suited for describing autonomous UAV systems that require both architectural clarity and traceability.
Although SysML has been applied in aerospace and UAV system modeling, a practical gap remains between SysML-based design models and executable robotics software architectures. Many UAV studies use SysML to represent requirements, block structures, and behaviors, but the mapping of these model elements to software components is often not explicitly defined. At the same time, ROS 2 is increasingly used for the implementation of robotic and autonomous systems because it supports modular nodes, publish-subscribe communication, request-response services, actions, lifecycle management, and distributed execution \cite{macenski2022,maruyama2016,erős2019}. However, ROS 2 development often starts directly from software implementation, with limited formal traceability back to stakeholder requirements and system-level design decisions. This creates a disconnect between formal systems engineering and robotics software development.

This paper addresses the above gap by proposing an MBSE design framework for the SysML-driven development of autonomous UAVs. The framework uses SysML as a formal design backbone and organizes UAV development across four connected layers: stakeholder requirements, functional decomposition, logical architecture, and physical/software allocation. At the requirements layer, stakeholder needs and mission objectives are captured and categorized into functional, performance, interface, safety, and compliance requirements. At the functional layer, autonomous UAV behaviors such as pre-flight checking, autonomous take-off, waypoint navigation, hover stabilization, obstacle avoidance, return-to-home, and landing are decomposed using activity and sequence models. At the logical architecture layer, subsystem responsibilities and interactions are defined using block definition diagrams, internal block diagrams, state machine diagrams, and parametric diagrams. At the physical/software allocation layer, SysML architectural elements are mapped to ROS 2 software concepts, including nodes, topics, services, and actions.

The main contributions of this paper are as follows. First, it proposes a four-layer MBSE design framework tailored to the development of autonomous UAVs. Second, it defines a SysML artifact set to capture the requirements, functions, logical architecture, interfaces, behaviors, and parametric constraints of an autonomous UAV system. Third, it provides a systematic mapping between SysML elements and ROS 2 software architecture elements, including the mapping of SysML blocks to ROS 2 nodes, flow ports and connectors to topics, request-response interactions to services, and goal-oriented behaviors to actions. Fourth, it establishes a design traceability and verification planning approach that prepares the UAV system for future Gazebo simulation, hardware-in-the-loop validation, and real-world flight testing.

The remainder of this paper is organized as follows. Section 2 reviews related work on MBSE, SysML, autonomous UAV system design, and ROS 2-based robotic software architectures. Section 3 presents the proposed MBSE-SysML-ROS 2 design framework. Section 4 describes the SysML-driven design of the autonomous UAV system, including requirements, functional, structural, behavioral, and parametric models. Section 5 presents the mapping from SysML architecture to ROS 2 software architecture and discusses design-phase verification planning. Section 6 discusses the implications, limitations, and future validation roadmap. Section 7 concludes the paper.

\section{Background and Related Work}

MBSE has become an important approach for developing complex cyber-physical systems because it enables requirements, architecture, behavior, interfaces, and verification information to be represented within a connected system model. In document-centric development, requirements specifications, interface descriptions, software design documents, and verification plans are often maintained separately. This separation can create ambiguity, inconsistency, and traceability gaps, especially when changes in one subsystem affect other parts of the system. For autonomous UAVs, these issues are critical because perception, flight control, navigation, communication, power monitoring, and mission management must operate as an integrated system under safety and timing constraints \cite{incose2023,iso15288,madni2018}.

The SysML is widely used to implement MBSE because it provides standard modeling constructs for requirements, structure, behavior, interfaces, and parametric constraints \cite{friedenthal2014,delligatti2013,omg2019,macenski2022,maruyama2016,erős2019,kemsaram2021mbse,kemsaram2021sysml}. Requirement diagrams can capture stakeholder and system requirements, while activity and sequence diagrams describe mission workflows and subsystem interactions. Block definition diagrams and internal block diagrams support architectural decomposition and interface specification, while state machine diagrams define operational modes such as take-off, waypoint navigation, hover, return-to-home, landing, and emergency handling. Parametric diagrams further support the representation of performance constraints such as navigation accuracy, communication latency, endurance, and obstacle detection range. Previous UAV-related studies have shown that SysML can improve architectural clarity and requirements traceability during system design \cite{hossain2022,wang2021,kemsaram2017design}.

ROS 2 is increasingly used for autonomous robotic and UAV software development because it supports modular nodes, publish-subscribe communication, request-response services, actions, lifecycle management, and distributed execution \cite{macenski2022}. These features make ROS 2 well-suited for implementing autonomous UAV functions, including mission management, navigation, perception processing, telemetry handling, and health monitoring. However, ROS 2 development often begins with software implementation, and the relationships among stakeholder requirements, system architecture, subsystem responsibilities, and software interfaces may not be formally defined. This can lead to interface drift, duplicated responsibilities, missing safety dependencies, and difficulty in verifying whether the implemented software satisfies the original system requirements.

Several SysML-based domain-specific approaches have addressed UAV and ROS-oriented robotic-system development. Aloui et al. proposed UavSwarmML, a domain-specific SysML model for representing UAV swarm missions, hardware configurations, and software implementation, followed by ROS-based simulation and validation \cite{aloui2022}. Winiarski introduced MeROS, a SysML-based metamodel covering ROS 1 and ROS 2 concepts for representing both the running robotic system and its development workspace \cite{winiarski2023}. Guizani et al. proposed ROS2ML, a SysML profile incorporating ROS 2 concepts to support continuity from autonomous mobile robot design to simulation and implementation \cite{guizani2023}. 
However, these existing approaches either focus on architecture modeling without a clear path to software allocation or on ROS 2 implementation without formal requirements traceability. A practical gap, therefore, remains between SysML-based system design and ROS 2-based software architecture for autonomous UAVs.

This paper addresses this gap by proposing a design-phase MBSE-SysML-ROS 2 framework for the development of autonomous UAVs. The framework connects stakeholder requirements, functional decomposition, logical architecture, and physical/software allocation. It defines how SysML artifacts can be used to model UAV requirements, mission behaviors, subsystem structures, interface flows, state logic, and performance constraints. It further maps SysML blocks, ports, connectors, and behaviors to ROS 2 nodes, topics, services, and actions. Unlike simulation-driven studies, this work focuses on establishing a traceable design baseline that prepares the UAV system for subsequent Gazebo simulation, hardware-in-the-loop validation, and real-world flight testing.

\section{Proposed MBSE-SysML-ROS 2 Design Framework}

\begin{figure}[!t]
\centering
\includegraphics[width=\textwidth]{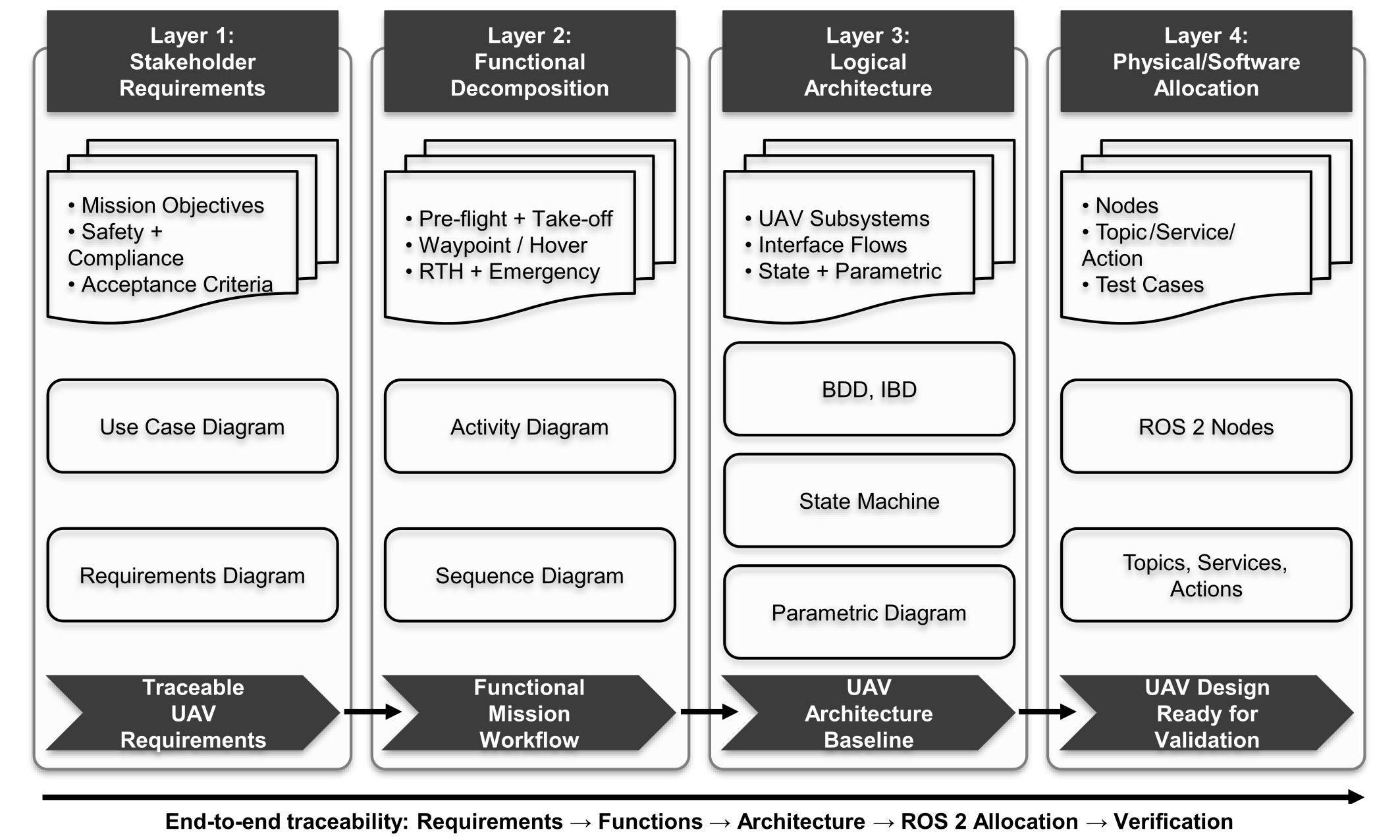}
\caption{
Proposed MBSE-SysML-ROS 2 design framework for autonomous UAV development. The framework connects stakeholder requirements, functional decomposition, logical architecture, and physical/software allocation to establish a traceable design baseline before simulation and physical deployment.
}
\label{fig:framework}
\end{figure}

The proposed framework provides a design-phase workflow for developing autonomous UAV systems using MBSE, SysML, and ROS 2. The main objective is to establish a traceable system design before simulation, hardware-in-the-loop validation, or real-world flight testing. The framework is organized into four connected layers: i) stakeholder requirements, ii) functional decomposition, iii) logical architecture, and iv) physical/software allocation. Each layer produces specific SysML artifacts and maintains traceability to the next layer.
The complete four-layer workflow of the proposed MBSE-SysML-ROS 2 design framework is summarized in Fig. \ref{fig:framework}.

The first layer captures stakeholder needs, mission objectives, operating assumptions, and system requirements. Requirements are grouped into functional, performance, interface, safety, and compliance categories. For an autonomous UAV, these requirements include capabilities such as autonomous take-off, waypoint navigation, hover stabilization, obstacle avoidance, return-to-home, telemetry exchange, power monitoring, and emergency handling. Each requirement is assigned an identifier, acceptance criterion, and verification method in line with requirements-engineering practice \cite{iso29148}. This supports early verification planning and prevents requirements from remaining as informal textual statements.

The second layer decomposes the UAV mission into functional behaviors. SysML activity and sequence diagrams are used to describe mission workflows and interactions among major system functions. Typical functions include mission upload, pre-flight checks, autonomous take-off, waypoint tracking, perception-based obstacle detection, path replanning, hover control, return-to-home, and landing. This layer provides a technology-independent view of what the UAV must do before assigning responsibilities to specific subsystems or software components.

The third layer defines the logical architecture of the autonomous UAV system. The architecture is represented using SysML block definition diagrams, internal block diagrams, state machine diagrams, and parametric diagrams. The UAV is decomposed into six major logical subsystems: Flight Control System, Navigation and Localization System, Perception and Sensing System, Communication System, Mission Management System, and Power Management System. The internal interfaces among these subsystems are defined through flow ports and connectors. The system-level state machine captures operating modes such as Idle, Pre-flight Check, Take-off, Navigate, Hover, Obstacle Avoidance, Return-to-Home, Landing, and Emergency. Parametric models are used to represent design constraints related to navigation accuracy, communication latency, endurance, and safety margins.

The fourth layer maps the logical architecture to a ROS 2-oriented software architecture. In this layer, SysML blocks are allocated to ROS 2 nodes, flow ports and connectors are mapped to topics, request-response interactions are mapped to services, and long-running goal-oriented behaviors are mapped to actions. For example, the Mission Management System can be mapped to an \texttt{mms\_node}, the Navigation and Localization System to an \texttt{nls\_node}, the Perception and Sensing System to a \texttt{pss\_node}, the Communication System to a \texttt{cs\_node}, and the Power Management System to a \texttt{pms\_node}. Continuous information, such as navigation state, obstacle data, mission state, telemetry, and battery status, is published as ROS 2 topics. Commands such as mission upload, mode change, and health-check requests are represented as services. Behaviors such as waypoint navigation and return-to-home are represented as actions.

\section{SysML-Driven Design of the Autonomous UAV System}

This section presents the design-level SysML model of the autonomous UAV system. The purpose is to define the system context, requirements, mission functions, logical architecture, interfaces, operating states, and design constraints before implementation or simulation. The model is structured to support traceability from stakeholder needs to ROS 2-oriented software allocation and future verification cases.

\subsection{System Context and Mission Scenarios}

The autonomous UAV system is designed for representative aerial missions, such as waypoint navigation, infrastructure inspection, obstacle avoidance, return-to-home, and emergency handling. The main external actors are the mission operator, ground control station, communication network, GNSS source, operating environment, and inspection target. The UAV receives mission commands and waypoints from the ground control station, estimates its state using onboard sensors, executes mission behaviors through the flight control system, and transmits telemetry and health information back to the operator.

Three representative design scenarios are considered. The first scenario is autonomous waypoint navigation, where the UAV performs pre-flight checks, takes off, follows a sequence of waypoints, stabilizes at required points, and returns home after mission completion. The second scenario is infrastructure inspection, where the UAV follows a planned path around a target structure and executes payload-related actions at predefined observation points. The third scenario is sense-and-avoid, in which the UAV detects an obstacle, enters an avoidance or hover state, replans its path, and resumes the mission when it is safe to do so. These scenarios provide the basis for deriving requirements, functions, state transitions, interfaces, and future verification tests.

\subsection{Requirements Modeling}

The requirements model is represented using SysML requirement diagrams. Requirements are organized into five categories: functional, performance, interface, safety, and compliance requirements. Functional requirements describe mission capabilities such as take-off, waypoint navigation, hover, obstacle avoidance, return-to-home, landing, and telemetry exchange. Performance requirements describe measurable design targets such as navigation accuracy, communication latency, update rate, endurance, and obstacle detection range. Interface requirements define information exchange among subsystems and with external actors. Safety requirements capture emergency behaviors such as low-battery response, link-loss handling, and fail-safe landing. Compliance requirements represent operational and regulatory constraints relevant to UAV deployment.

\subsection{Functional and Behavioral Modeling}

The functional model is captured using activity and sequence diagrams. The activity model decomposes the mission into major functions: mission upload, pre-flight health check, system arming, take-off, waypoint tracking, hover stabilization, payload operation, obstacle monitoring, return-to-home, and landing. Decision nodes represent mission conditions such as an obstacle detected, a waypoint reached, a low battery, communication lost, or the mission completed.

Sequence diagrams are used to describe the time-ordered interactions among the ground control station, mission management system, navigation system, perception system, flight control system, communication system, and power management system. For example, in the mission initiation sequence, the ground control station uploads the mission plan, the mission management system validates it, the navigation system confirms localization readiness, the power management system confirms battery status, and the flight control system receives the take-off command. These diagrams help identify required interfaces before ROS 2 implementation.


The system-level behavior is represented using a SysML state machine diagram with the following operating modes: Idle, Pre-flight Check, Take-off, Navigate, Hover, Obstacle Avoidance, Return-to-Home, Landing, and Emergency. Transitions are defined using triggers and guard conditions. For example, Navigate can transition to Return-to-Home after mission completion, low-battery detection, link loss, or operator command, while Navigate can transition to Obstacle Avoidance when an obstacle is detected. This state-based model serves as the design basis for future ROS 2 mission-management logic.

\begin{figure}[!t]
\centering
\includegraphics[width=\textwidth]{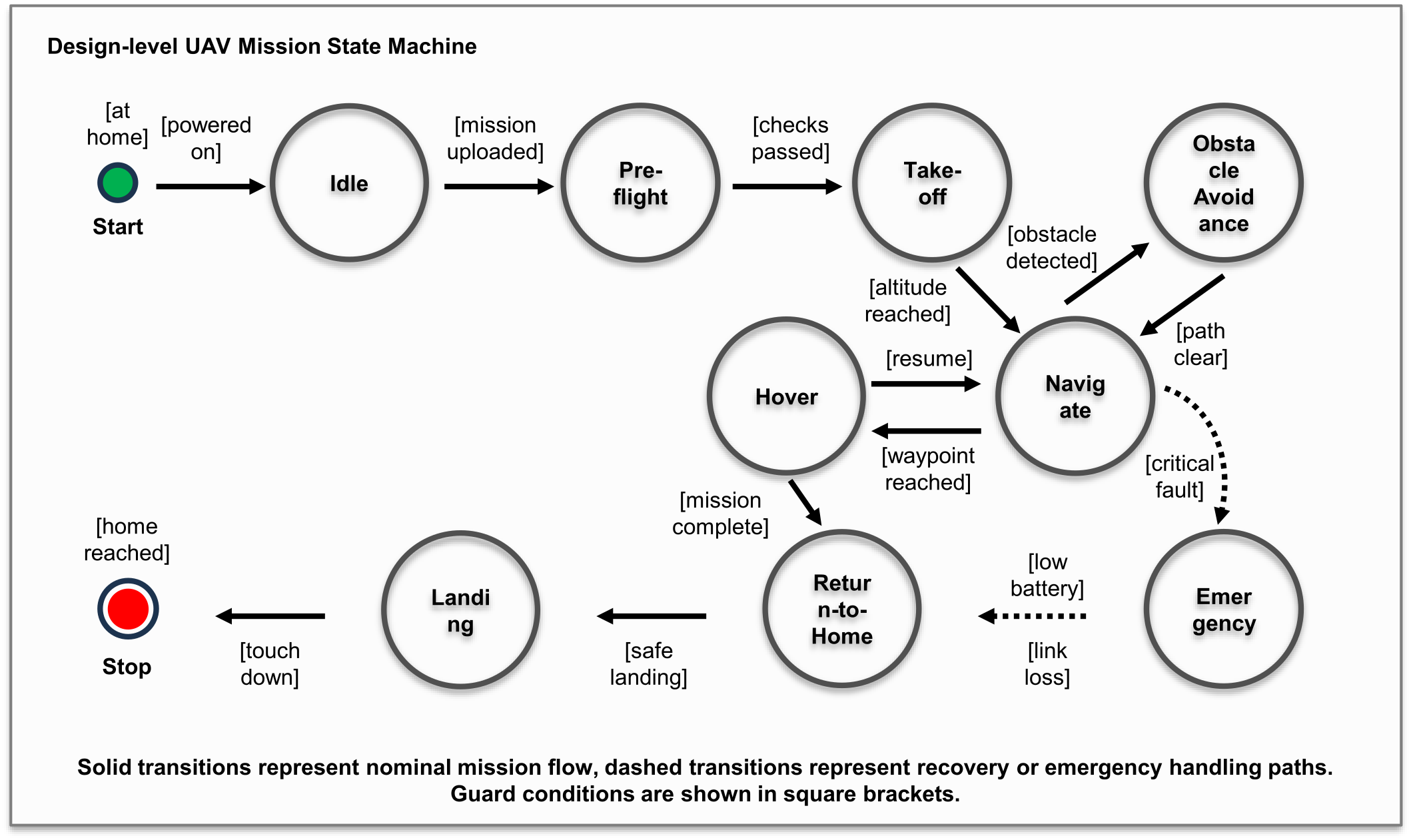}
\caption{
Mission state machine for autonomous UAV operation. The design-level state machine defines the main UAV operating modes, including Idle, Pre-flight Check, Take-off, Navigate, Hover, Obstacle Avoidance, Return-to-Home, Landing, and Emergency. The transitions capture nominal mission progression and safety-related recovery paths triggered by obstacle detection, mission completion, low battery, link loss, or critical fault conditions.
}
\label{fig:mission_state_machine}
\end{figure}

The design-level mission state machine is shown in Fig. \ref{fig:mission_state_machine}, where nominal mission transitions and safety-related recovery paths are defined before ROS 2 mission-management implementation.

\subsection{Logical Architecture and Interface Modeling}

\begin{figure}[!t]
\centering
\includegraphics[width=\textwidth]{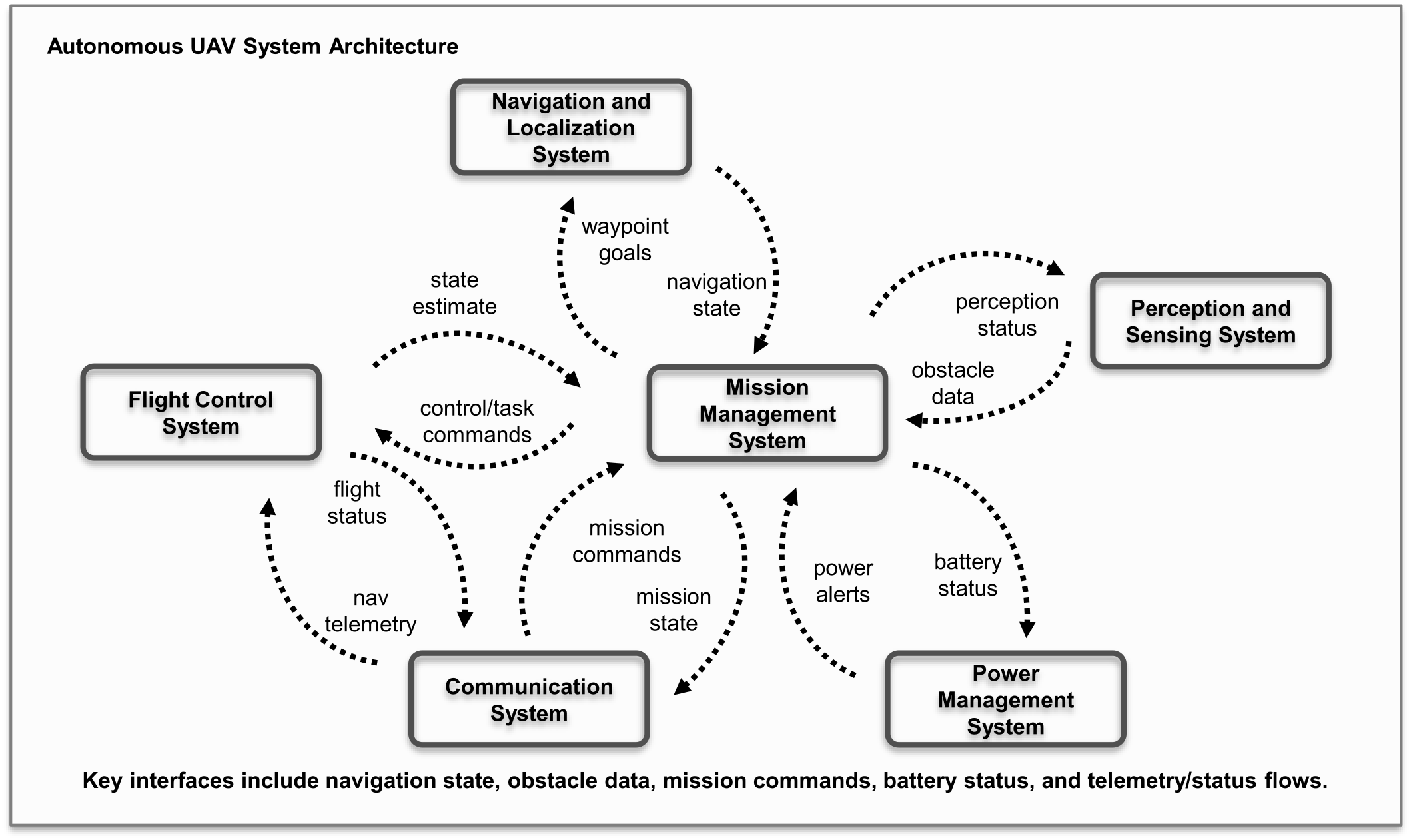}
\caption{
SysML logical architecture of the autonomous UAV system. The design decomposes the UAV into six major subsystems: Flight Control System, Navigation and Localization System, Perception and Sensing System, Communication System, Mission Management System, and Power Management System. The main data and command flows provide the basis for later ROS 2 topic, service, and action mapping.
}
\label{fig:sysml_logical_architecture}
\end{figure}

The logical architecture is represented using SysML block definition diagrams and internal block diagrams. The UAV system is decomposed into six major subsystems: Flight Control System, Navigation and Localization System, Perception and Sensing System, Communication System, Mission Management System, and Power Management System. Each subsystem has a clearly defined design responsibility. The Flight Control System manages low-level stabilization and actuator commands. The Navigation and Localization System estimates UAV state and supports waypoint tracking. The Perception and Sensing System detects obstacles and provides environmental information. The Communication System manages mission commands and telemetry. The Mission Management System coordinates mission logic and state transitions. The Power Management System monitors battery status and provides energy-related alerts.
The proposed SysML logical architecture is shown in Fig. \ref{fig:sysml_logical_architecture}, where subsystem responsibilities and major interface flows are defined before ROS 2 software allocation.

The internal block diagram defines the data and command flows among these subsystems. Navigation state flows from the Navigation and Localization System to the Flight Control and Mission Management systems. Obstacle information flows from the Perception and Sensing System to the Mission Management System. Battery status flows from the Power Management System to the Mission Management and Communication systems. Mission commands flow from the Communication System to the Mission Management System, then to the relevant subsystems. These interface definitions form the basis for later mapping to ROS 2 topics, services, and actions.

\subsection{Parametric and Verification Planning Model}

Parametric diagrams are used to represent design constraints that will guide later simulation and physical validation. The main parametric concerns include navigation accuracy, communication latency, endurance, obstacle detection range, and return-to-home response time. At the design phase, these constraints are used to define acceptance criteria rather than to report completed performance results.
The verification planning model links requirements to future test cases using SysML verify relationships. For example, waypoint navigation requirements are linked to future Gazebo mission tests, communication-link-loss requirements are linked to hardware-in-the-loop and simulation tests, and low-battery return-to-home requirements are linked to power-management validation tests. This approach ensures that the design model is ready for systematic validation once simulation and hardware testing become available.

The system models were specified using OMG SysML v2 and developed using the CATIA Magic/Cameo SysML v2 Community Edition provided by Dassault Systèmes. The target software environment for the planned implementation and validation is ROS 2 Jazzy Jalisco running on Ubuntu 24.04 LTS. As the present study focuses on system design, this ROS 2 environment represents the intended platform for subsequent Gazebo simulation and implementation rather than a completed software deployment.

\section{ROS 2 Software Architecture Mapping and Verification}

This section describes how the SysML logical architecture is mapped to a ROS 2-oriented software architecture at the design phase. The purpose is to define the intended software structure and communication interfaces before implementation, simulation, or physical testing. This mapping provides a bridge between the SysML system model and the future ROS 2 implementation.
The proposed software architecture targets ROS 2 Jazzy Jalisco on Ubuntu 24.04 LTS.

\subsection{SysML-to-ROS 2 Mapping Rules}

The mapping follows four main rules. First, each major SysML subsystem block is allocated to a corresponding ROS 2 node. Second, SysML flow ports and internal block diagram connectors are mapped to ROS 2 topics when they represent continuous data streams. Third, discrete command-and-query interactions are mapped to ROS 2 services. Fourth, long-running goal-oriented behaviors are mapped to ROS 2 actions. 
The proposed mapping is informed by previous ROS-oriented SysML metamodels and profiles, particularly MeROS and ROS2ML \cite{winiarski2023,guizani2023}. However, the mapping rules in this study are reformulated using SysML v2 model elements and tailored to autonomous UAV architecture and verification planning.

\subsection{Proposed ROS 2 Node Architecture}

\begin{figure}[!t]
\centering
\includegraphics[width=\textwidth]{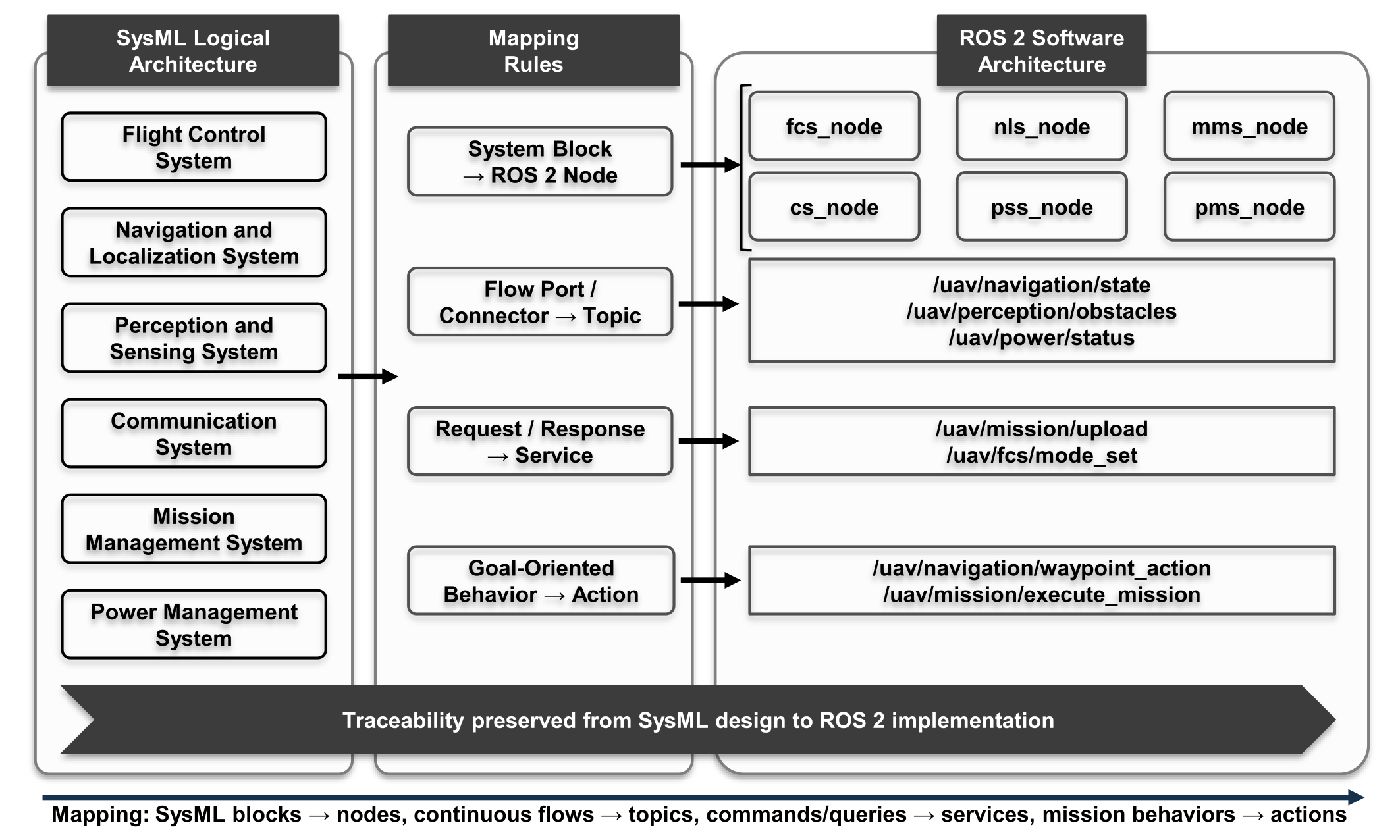}
\caption{
SysML-to-ROS 2 software architecture mapping. SysML subsystem blocks are allocated to corresponding ROS 2 nodes, interface flows are mapped to topics, command-and-query interactions are mapped to services, and long-running mission behaviors are mapped to actions. This mapping preserves traceability between the SysML logical architecture and the planned ROS 2 software design.
}
\label{fig:sysml_ros2_mapping}
\end{figure}

The proposed ROS 2 architecture consists of six primary nodes derived from the SysML logical architecture. The \textit{fcs\_node} represents the Flight Control System and interfaces with the autopilot layer for stabilization and actuator-level control. The \textit{nls\_node} represents the Navigation and Localization System and is responsible for state estimation, waypoint tracking, and navigation-state reporting. The \textit{pss\_node} represents the Perception and Sensing System and provides obstacle and environment information. The \textit{cs\_node} represents the Communication System and manages telemetry, command reception, and link-status monitoring. The \textit{mms\_node} represents the Mission Management System and coordinates mission sequencing, decision-making, and state transitions. The \textit{pms\_node} represents the Power Management System and publishes battery status, low-power warnings, and power-related alerts.
The proposed mapping from SysML logical architecture elements to ROS 2 software components is shown in Fig. \ref{fig:sysml_ros2_mapping}.

The proposed communication architecture includes topics for continuous information exchange, services for mission and system commands, and actions for long-running autonomous behaviors. Navigation state, obstacle information, mission state, telemetry, link status, and battery status are represented as topics. Mission upload, mode setting, system health query, and emergency command execution are represented as services. Waypoint navigation, mission execution, and return-to-home are represented as actions. This communication design ensures that each ROS 2 interface can be traced back to a corresponding SysML connector, flow port, requirement, or behavioral model.

\section{Discussion, Limitations, and Future Work}

The proposed MBSE–SysML–ROS 2 framework provides a structured design approach for autonomous UAV development by connecting system requirements, functional behaviors, logical architecture, software allocation, and verification planning within a unified model. In this study, the system architecture is specified using SysML v2 and developed with the Dassault Systèmes CATIA Magic/Cameo SysML v2 Community Edition. The planned implementation environment targets ROS 2 Jazzy Jalisco on Ubuntu 24.04 LTS.

The framework establishes a conceptual mapping between SysML v2 model elements and ROS 2 software components. SysML v2 part definitions and part usages represent UAV subsystems and are allocated to corresponding ROS 2 nodes. Ports, connections, and item flows represent information exchange. They are mapped to ROS 2 topics, while discrete request–response interactions and long-running mission behaviors are mapped to ROS 2 services and actions, respectively. Requirement, allocation, satisfaction, and verification relationships provide traceability from stakeholder needs to system functions, architectural elements, ROS 2 interfaces, and planned test cases.

The current mapping is design-oriented and manually maintained within the modeling workflow. Therefore, the framework does not presently provide automatic code generation or fully automated bi-directional synchronization between the SysML v2 model and the ROS 2 implementation. Changes to requirements, architectural elements, or ROS 2 interfaces must be reviewed and propagated through the corresponding traceability links and allocation records. This limitation will be addressed in future work through model transformation rules, machine-readable interface specifications, and automated consistency checking between SysML v2 and ROS 2 artifacts.

Future work will implement the proposed architecture using ROS 2 Jazzy Jalisco, PX4, and Gazebo simulation \cite{koenig2004} on Ubuntu 24.04 LTS, followed by hardware-in-the-loop and real-world flight experiments \cite{meier2015}. The framework will also be extended to cooperative multi-UAV systems and evaluated for scalability, traceability maintenance, interface consistency, and development effort.

\section{Conclusion}

This paper presented a design-phase MBSE-SysML-ROS 2 framework for the structured development of autonomous UAV systems. The proposed framework uses SysML as a formal design backbone to connect stakeholder requirements, functional decomposition, logical architecture, physical/software allocation, and verification planning. By mapping SysML blocks, interfaces, behaviors, and verification cases to ROS 2 nodes, topics, services, actions, and future test scenarios, the framework provides a traceable bridge between system-level design and robotics software architecture. 
It establishes a reproducible design baseline that reduces early-stage ambiguity, improves interface consistency, and prepares the autonomous UAV system for future Gazebo simulation, hardware-in-the-loop validation, and real-world flight testing.

\end{document}